\documentclass[10pt,journal,compsoc]{IEEEtran}
\ifCLASSOPTIONcompsoc
  \usepackage[nocompress]{cite}
\else
  \usepackage{cite}
\fi

\ifCLASSINFOpdf
\else
\usepackage[dvips]{graphicx}
\graphicspath{{./figs/}}
\DeclareGraphicsExtensions{.eps}
\fi

\usepackage{amsmath}
\usepackage{algorithmic}
\usepackage{bm}
\usepackage{amssymb}
\usepackage{color}
\newcommand{\argmin}{\operatornamewithlimits{argmin}}
\newcommand{\argmax}{\operatornamewithlimits{argmax}}

\begin{document}
%\title{Bare Advanced Demo of IEEEtran.cls for\\ IEEE Computer Society Journals}
\title{Structural Data Recognition with Graph Model Boosting}

\author{Tomo~Miyazaki%,%~\IEEEmembership{Member,~IEEE,}
  ~and~Shinichiro~Omachi%,%~\IEEEmembership{Senior~Member,~IEEE}%
  %and~Jane~Doe,~\IEEEmembership{Life~Fellow,~IEEE}% <-this % stops a space
  \IEEEcompsocitemizethanks{\IEEEcompsocthanksitem T. Miyazaki and S. Omachi are
    with the Graduate School of Engineering, Tohoku University, Japan.\protect\\
    %E-mail: see http://www.michaelshell.org/contact.html
    E-mail: tomo@iic.ecei.tohoku.ac.jp, machi@ecei.tohoku.ac.jp
    %\IEEEcompsocthanksitem J. Doe and J. Doe are with Anonymous University.
  }% <-this % stops a space
}

\IEEEtitleabstractindextext{%
  \begin{abstract}
    This paper presents a novel method for structural data recognition
    using a large number of graph models.
    In general, prevalent methods for structural data recognition have two shortcomings:
    1) Only a single model is used to capture structural variation.
    2) Naive recognition methods are used, such as the nearest neighbor method. 
    In this paper, we propose strengthening the recognition performance of these models
    as well as their ability to capture structural variation.
    The proposed method constructs a large number of graph models
    and trains decision trees using the models.
    This paper makes two main contributions.
    The first is a novel graph model that can quickly perform calculations,
    which allows us to construct several models in a feasible amount of time.
    The second contribution is a novel approach to structural data recognition: graph model boosting.
    Comprehensive structural variations can be captured with a large number of graph models
    constructed in a boosting framework,
    and a sophisticated classifier can be formed by aggregating the decision trees.
    Consequently, we can carry out structural data recognition with powerful recognition capability
    in the face of comprehensive structural variation.
    The experiments shows that the proposed method achieves impressive results and outperforms existing methods on datasets of IAM graph database repository. 
  \end{abstract}

%\begin{IEEEkeywords}
%Computer Society, IEEE, IEEEtran, journal, \LaTeX, paper, template.
  %\end{IEEEkeywords}
}

\maketitle

% is not selected <OR> if conference mode is selected - because compsoc
% conference papers position the abstract like regular (non-compsoc)
\IEEEdisplaynontitleabstractindextext

% \fi
\IEEEpeerreviewmaketitle

\ifCLASSOPTIONcompsoc
\IEEEraisesectionheading{\section{Introduction}\label{sec:introduction}}
\else
\section{Introduction}
\label{sec:introduction}
\fi
\IEEEPARstart{S}{tructural} data represented by graphs
are a general and powerful representation of objects and concepts.
A molecule of water can be represented as a graph with three vertices and an edge,
where the vertices represent hydrogen and oxygen,
and the relation is intuitively described by the edge. 
Structural data recognition is used in a wide range of applications; 
for example, in bioinfomatics, chemical compounds need to be recognized as active or inactive.

The difficulty of graph recognition arises out of a lack of mathematical analysis.
Even measuring the distance between graphs requires various techniques.
The problem of graph recognition has recently been actively studied \cite{White:2008:ICPR, Han:2012:SSPR, Zhang:2010:ICPR, Riesen:2009:GbR, Bunke:2012:PRL}.
Past research has led to notable progress in two aspects.
First, graph models have been developed to capture structural variations.
Second, the embedding of graphs into Euclidean space has been used
to apply sophisticated recognition rules to the vector domain.
However, both these aspects have drawbacks.
The drawback of the former is that only naive recognition rules are applicable,
such as the nearest neighbor (NN) or the k-nearest neighbor (k-NN) methods,
which require a considerable amount of training data to achieve high performance.
It is unusual for a sufficient amount of data to be available.
The drawback of the second aspects above is the loss of structural variation in the embedding process.

Our aim in this paper is to overcome the drawbacks of the previous methods. Specifically, we develop a novel graph model and a boosting framework for structural data to strengthen recognition capability and capture structural variation. These are the two main contributions of this paper.
We introduce each briefly below,
and then describe them in greater detail in Sections \ref{sec:Model} and \ref{sec:Boosting}.

The first contribution of this paper is a method to construct a novel graph model. Our model captures four types of structural variations, in vertex composition, edge composition, vertex attributes, and edge attributes.  We need vertex correspondence among graphs for model construction, but it is challenging to obtain such correspondences. Most existing methods consider correspondence between graphs \cite{Cho:2010:ECCV, Cho:2013:ICCV, Cour:2007:NIPS, Gold:1996:PAMI}. We propose searching vertex correspondences among multiple graphs by assigning labels to vertices with median graphs \cite{Jiang:2001:PAMI}. The calculation of the median graph is relatively easy. This property helps us obtain the correspondences in a feasible amount of time. This is an important property, since we construct a large number of graph models in the proposed method.
Moreover, we develop similarity between a graph and our graph model. 

The second contribution of this paper is a novel approach to capture structural variation. In order to capture structural variation comprehensively, we construct a large number of models so that they can contain different structural variations and compensate one another.  We then use a boosting method and the models to construct a classifier.  Consequently, we can equip the classifier with comprehensive structural variation and a powerful recognition capability.  We call this novel approach {\it graph model boosting} (GMB). 

In experiments, we demonstrated structural data recognition
using GMB on five graph datasets that were publicly available.
We confirmed that accuracy of GMB notably increased as the boosting process continued. 
GMB also outperformed existing methods in the area by a large margin.
The experimental results thus show the effectiveness of the GMB. 

A preliminary version of the work reported here was first presented in a conference paper \cite{Miyazaki:2016:ICPR}.
We consolidate and expand our previous description and results. Firstly, we provide additional technical details concerning the graph model and GMB. Our contributions are highlighted clearly in the Abstract and Introduction.  Secondly, we carried out a wider survey about related work to clarify the significance of the proposed method.  Lastly, additional experimental results are presented: time complexity to evaluate practicality,
the impact of the parameters for robustness assessment to various datasets, and using other graph model to show the capability of GMB.

\section{Related Work} \label{sec:related}
Existing methods for graph recognition can be broadly categorized into three approaches:
the one-vs-one approach, the model-based approach, and the embedding approach. 

Methods in the one-vs-one approach attempt to classify graphs according to a criterion that can be measured for two graphs, such as graph isomorphism, subgraph isomorphism, and graph edit distance. Two graphs are graph isomorphic if the mapping of their vertices and edges is bijective. Subgraph isomorphism is the case where subgraphs of two graphs satisfy graph isomorphism. Isomorphism is determined by tree search \cite{Ullmann:1976} or
backtracking \cite{Ghahraman:1980}. However, it is difficult to determine subgraph isomorphism in case of noisy vertices and edges. Therefore, methods using graph edit distance have been developed. 
The graph edit distance is defined as the minimum sum of the costs of edit operations that transform the source graph into the target graph by substitution, deletion, and insertion \cite{Eshera:1984, Sanfeliu:1983}. Since it is expensive to search every combination of edit operations, approximate edit operations are searched. The method proposed in \cite{Neuhaus:2006:SSPR} applies the A-star algorithm \cite{Hart:1968:Astar} and the one in \cite{Riesen:2009:GbR} exploits Munkres's algorithm \cite{Munkres:1957}.  The literature \cite{Conte:2004, Bille:2005} will help readers find more details related to graph matching and edit distance. The advantage of this approach is that the calculation is completed in polynomial time. However, the methods in this approach only adopt naive recognition rules, such as the NN method and the k-NN method. 
Furthermore, they only consider local structural variations
because the edit distance is measured between two graphs; other graphs are ignored.

Methods in the model-based approach attempt to capture structural variation
and classify graphs using a model.
The median graph \cite{Jiang:2001:PAMI} is a model
that captures global information concerning graphs.
The median graph minimizes the total distance between training graphs.
A random graph \cite{Wong:1985:PAMI} is a model
that specializes in capturing attribute variations at each vertex and edge.
The random graph contains variables ranging from 0 to 1,
which are associated with attribute values.
The variables represent the probabilities of the attribute that the vertices take.
However, numerous variables are required when attributes are continuous values.
Improved models \cite{Bagdanov:2003:PR, Serratosa:2003:PR, SanFeLiu:2004} based on random graph have been developed as well. There are three such models: First-order Gaussian graph, or FOGG \cite{Bagdanov:2003:PR}, function-described graph, or FDG \cite{Serratosa:2003:PR}, and second-order random graphs, or SORGs \cite{SanFeLiu:2004}. The FOGG is a model designed to avoid increasing number of variables by replacing those of a random graph with parameters of a Gaussian distribution. FDG and SORG incorporate joint probabilities among the vertices and the edges to describe structural information. The difference between FDG and SORG is the numbers of vertices and edges at the calculation of joint probability. FDG uses pairs of vertices or edges, whereas multiple vertices and edges are used in SORG. Recently, models exploiting unsupervised learning methods have been developed \cite{Torsello:2006:PAMI}. Torsello and Hancock presented a framework to integrate tree graphs into one model by minimizing the minimum description length \cite{Rissanen:1978}. Furthermore, Torsello \cite{Torsello:2008:CVPR} expanded tree graphs to graphs and adopted a sampling strategy \cite{Hammersley:1964} to improve calculation efficiency. The EM algorithm has been applied to construct a model \cite{Han:2010:ICPR}. The methods in \cite{Torsello:2006:PAMI, Torsello:2008:CVPR, Han:2010:ICPR} concentrate on capturing variations in vertex and edge composition.  For computational efficiency, a closure tree \cite{He:2006} has been developed. Each vertex of the tree contains information concerning its descendants, so that effective pruning can be carried out. The model-based approach can measure distance based on structural variation. However, these methods also adopt naive recognition rules, the same as in the one-vs-one approach.

Methods in the embedding approach attempt to apply sophisticated recognition rules
that are widely used in the vector domain.
The main obstacle to embedding graphs is
the lack of a straightforward and unique transformation from a graph to a vector.
For example, a graph with $N$ vertices can be transformed into $N!$ adjacency matrices,
since there are $N!$ permutations of vertex labeling. Jain and Wysotzki embedded graphs into Euclidean space using the Schur--Hadamard inner product and performed k-means clustering by utilizing neural networks \cite{Jain:2004:ML}. Bunke and Riesen \cite{Bunke:2012:PRL} used the graph edit algorithm to embed graphs into a vector space
and applied a support vector machine
\cite{Vapnik:1998}.
Some methods \cite{Kudo:2004:NIPS, Nowozin:2007:CVPR, Zhang:2010:ICPR} involve applying AdaBoost to graphs. 
We categorize them as part of this approach because they use sophisticated recognition methods. 
Kudo et al. developed a decision stump that responds to whether a query graph includes a specific subgraph 
and constructs a classifier using stumps in AdaBoost \cite{Kudo:2004:NIPS}. 
Nowozin et al. developed a classifier based on LPBoost \cite{Nowozin:2007:CVPR}. 
Zhang et al. \cite{Zhang:2010:ICPR} improved classifiers by incorporating an error-correcting coding matrix method \cite{Dietterich:1995} into boosting. 
Both \cite{Nowozin:2007:CVPR} and \cite{Zhang:2010:ICPR} adopted the decision stump as a weak classifier. 
The difference between GMB and these methods \cite{Kudo:2004:NIPS, Nowozin:2007:CVPR, Zhang:2010:ICPR} is the structural variation in the classifiers.
Comprehensive structural variation is incorporated into the GMB, whereas
local structural variation is used in \cite{Kudo:2004:NIPS, Nowozin:2007:CVPR, Zhang:2010:ICPR} because of the subgraphs.
The advantage of the embedding approach is that it can exploit powerful recognition methods.
However, a disadvantage is that it ignores structural variation,
since the graphs cannot be recovered from the vectors.
In addition, embedding graphs with structural variation is a challenging task.

Summarizing the one-vs-one approach and the model-based approach requires a powerful recognition method. 
The methods in the embedding approach need to incorporate structural variations. 
The advantages of the model-based and the embedding approaches can complement each other
to mitigate their disadvantages. 
Therefore, our strategy for overcoming the disadvantages is
to integrate the model-based approach with a sophisticated recognition method. 
In this paper, we build several graph models and 
incorporate them into AdaBoost to construct a graph recognizer
that can consider comprehensive structural variation.

\section{Graph Model} \label{sec:Model}
\subsection{Definition of Graph Model}
We propose a graph model that captures four types of structural variation in graphs:
vertex composition, edge composition, vertex attributes, and edge attributes.
The graph model $P$ is defined as 
\begin{equation}
  P = ( V, E, B, \Theta ), \label{eq:Model}
\end{equation}
where $V$, $E$, $B$, and $\Theta$ are the sets of vertices, edges,
the probabilities of the vertices and edges,
and the parameters of a probability density function of
the vertices and edges for attributes, respectively.
The compositions and attribute variations are captured by $B$ and $\Theta$, respectively.
We use probability density function $f_{\text{pdf}}$ to calculate the probability
that vertex $v_i$ takes attribute $a$ as $f_{\text{pdf}}(a|\theta_i)$.
We give an example of how the model describes variation in vertex composition.
Let $\{v_1, v_2, v_3\}$ and $\{b_1, b_2, b_3\}$ be elements of $V$ and $B$, respectively.
The vertex compositions and probabilities are as follows: $\{v_1\}$ at $b_1(1-b_2)(1-b_3)$,
$\{v_2\}$ at $(1-b_1)b_2(1-b_3)$, $\{v_3\}$ at $(1-b_1)(1-b_2)b_3$,
$\{v_1,v_2\}$ at $b_1b_2(1-b_3)$,
$\{v_1,v_3\}$ at $b_1(1-b_2)b_3$, $\{v_2,v_3\}$ at $(1-b_1)b_2b_3$,
and $\{v_1,v_2,v_3\}$ at $b_1b_2b_3$.
We can include attribute variations by multiplying $f_{\text{pdf}}(a|\theta)$.

We define likelihood function $f_L$
whereby a model generates an attributed graph $G'=(V',E',A')$.
Let $Y \in \{0,1\}^{|V|\times|V'|}$ be a matching matrix between $P$ and $G'$,
where $Y$ is subject to $\sum_{i=1}^{|V|} Y_{ik} \le 1$
and $\sum_{k=1}^{|V'|} Y_{ik} \le 1$ for any vertex $v'_k$ in $V'$ and $v_i$ in $V$.
If two vertices $v_i$ and $v'_k$ are matched, $Y_{ik}=1$, $Y_{ik}=0$ otherwise.
Therefore, $Y$ provides the correspondence between $P$ and $G'$.
We use function $\pi$ to refer to the corresponding vertex
$\pi: v \mbox{ in } V \to v' \mbox{ in } V'$.
We calculate the likelihood of $G'$ as 
\begin{eqnarray}
  f_L( G' , P ) = \max_Y \prod_{ v_i \in V_{\text{match}} }
  b_i f_{\text{pdf}}( a'_{\pi(v_i)} \mid \theta_i )
  \prod_{v_{\bar{i}} \in V_{\text{miss}} } ( 1 - b_{\bar{i}} ) \nonumber \\ 
  \prod_{e_{ij} \in E_{\text{match}} } b_{ij}
  f_{\text{pdf}}( a'_{\pi(v_i)\pi(v_j)} \mid \theta_{ij} )
  \prod_{ e_{\bar{i}\bar{j}} \in E_{\text{miss}}} ( 1 - b_{\bar{i}\bar{j}} ), \label{eq:prob}
\end{eqnarray}
\begin{eqnarray}
  V_{\text{match}} &=& \{ v_i \mid v_i \in V, v'_k \in V', Y_{ik}=1\}, \\ 
  V_{\text{miss}} &=& V \setminus V_{\text{match}}, \\
  E_{\text{match}} &=& \{ e_{ij} \mid e_{ij} \in E, e'_{kl} \in E', Y_{ik}=Y_{jl}=1 \}, \\
  E_{\text{miss}} &=& E \setminus E_{\text{match}}.
\end{eqnarray}
We call this graph model {\it probabilistic attribute graph gneration model, PAGGM}.

\subsection{Model Construction}\label{sec:model:construction}
The construction of $P$ involves the use of training data
  $\bm{G} = \{G_1, \cdots G_n\}$ and correspondences
  $\bm{Y} = \{Y^1, \cdots, Y^n\}$
between $P$ and $\bm{G}$.
Specifically,
we calculate $b_i$ and $\theta_i$ from vertices and attributes that correspond using $\bm{Y}$. 
We calculate $B$ and $\Theta$ of the edges likewise. 

Searching $\bm{Y}$ is a difficult problem.
Attempts have been made to estimate $\bm{Y}$ by minimizing objective functions,
such as minimum description length \cite{Torsello:2006:PAMI, Han-2015-pami}
and entropy \cite{Bunke:2003:GbPR}.
However, they are not applicable to our study
because we search $\bm{Y}$ for over thousands of subsets of $\bm{G}$.

We propose a method to quickly calculate the correspondence.
The key idea is to convert searching for $\bm{Y}$ into labeling vertices,
where vertices in $\bm{G}$ corresponding to the same vertex in $P$ share the same label.
We exploit median graphs for labeling.
The procedure for the assignment of labels is illustrated in Algorithm 1.
$\bm{M}$ is used for calculating correspondences between $P$ and a graph.
Our proposed method can provide correspondences in feasible time
because the calculation of the median graph takes $O(n^2)$ time,
where $n$ is the number of graphs in $\bm{G}$.
We calculate median graph $M$ as 
\begin{equation}
  M = \argmin_{G \in \bm{G}} \sum_{G_i \in \bm{G}} f_d( G, G_i ), \label{eq:median} 
\end{equation}
where $f_d$ is a function used to calculate the distance between graphs.
We adopt an edit distance algorithm \cite{Neuhaus:2006:SSPR} as $f_d$ in this paper.

%\begin{figure}[t]
  %\caption{Vertex label assigning} \label{alg:group} % Algorithm 1
  %\begin{algorithmic}[1]
  %\STATE Assign null label to every vertex of all graphs in $\bm{G}$.
  %\STATE Calculate the median graph $M$ of $\bm{G}$.
  %\STATE Assign a new label to each null-labeled vertex of $M$.
  %\STATE Calculate matching matrices between $M$ and all graphs in $\bm{G}$.
  %\STATE Assign vertex labels of $M$ to
  %  null-labeled vertices of graphs in $\bm{G}$ according to the matching matrices.
  %\STATE Remove a graph from $\bm{G}$ if all of its vertices have labels.
  %\STATE Put $M$ into a set of median graphs $\bm{M}$.
  %\STATE Stop if $\bm{G}$ is empty; otherwise, go to Step 2.
  %\end{algorithmic}
%\end{figure}

\begin{figure}[t] \centering
  \includegraphics{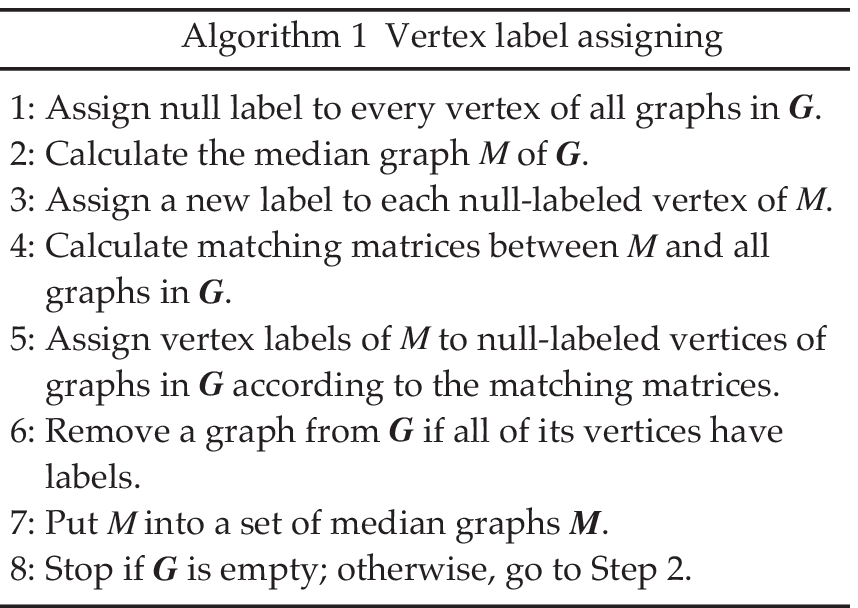}
\end{figure}

We create as many vertices and edges of PAGGM 
as the number of labels assigned by Algorithm 1.
It is straightforward to organize the labels of edges in $\bm{G}$
when the vertex labels of $\bm{G}$ are available.
Edges in $\bm{G}$ have the same label if the connected vertices of the edges have the same label.
Let $\mathcal{L}$ be a function that refers to the labels of vertices;
two edges $e_{1,2}$ and $e_{3,4}$ belong to the same labels
if $\{ \mathcal{L}(v_1), \mathcal{L}(v_2) \}$ is equal to $\{ \mathcal{L}(v_3), \mathcal{L}(v_4) \}$. 

We calculate $b_i$ and $b_{ij}$ in $B$ using the labels as 
\begin{eqnarray}
  b_i &=& \frac{1}{|\bm{G}|} \sum_{G_k \in \bm{G}} \sum_{v \in V_k}
  \mathbb{I}\left( \mathcal{L}(v)=i \right), \\
  b_{ij} &=& \frac{1}{|\bm{G}|} \sum_{G_k \in \bm{G}} \sum_{e_{v,v'} \in E_k}
  \mathbb{I}\left( \{ \mathcal{L}(v), \mathcal{L}(v') \} = \{i,j\} \right), 
\end{eqnarray}
where $\mathbb{I}$ is a function of proposition $p$ and takes $1$ if $p$ is true,
and otherwise takes $0$.
We can accumulate the attributes of vertex $v_i$ and edge $e_{ij}$ in $P$ as 
\begin{eqnarray}
  A_i &=& \bigcup_{G_k \in \bm{G} } \bigcup_{\substack{v_l \in V_k\\ \mathcal{L}(v_l)=i}} a_l,\\
  A_{ij} &=& \bigcup_{G_k \in \bm{G} }
  \bigcup_{\substack{e_{l,m} \in E_k\\ \{\mathcal{L}(v_l),\mathcal{L}(v_m)\} = \{i,j\}}} a_{l,m}.
\end{eqnarray}
Then, we calculate $\Theta$ using $A_i$ and $A_{ij}$.
However, we need to determine the type of the probability density function before calculating $\Theta$.
There is a miscellany of types of attributes,
such as continuous or discrete values, finite or infinite set, etc.
A suitable function is determined by a type of attribute.
In this paper, we adopt a normal density function and a discrete density function
for continuous and discrete values, respectively.

\subsection{Similarity} 
We describe similarity between graphs and the PAGGM. An intuitive choice for similarity is the logarithm of Eq. (\ref{eq:prob}). This choice is favorable when every vertex and edge in $G'$ matches that in $P$. However, unmatched vertices and edges in $G'$ occur often. In such cases, the logarithm of Eq. (\ref{eq:prob}) is unsuitable because the unmatched vertices and edges in $G'$ are not accounted for in Eq. (\ref{eq:prob}). Therefore, we impose penalty $\eta$ on unmatched vertices and edges. We define similarity function $f_s$ of graph $G'$ to a PAGGM $P$ as  

\begin{equation}
  f_s(G',P) = \log \left( f_L(G',P) \right) -
  \eta \left( |\bar{V'}| + |\bar{E'}| \right), \label{eq:similarity}
\end{equation}
where $\bar{V'}$ and $\bar{E'}$ represent sets of unmatched vertices and edges of $G'$, respectively. We experimentally set $\eta$ to $3$ in this paper. 

\begin{figure}[t] \centering
  \begin{tabular}{cc}
    \scalebox{1}[1]{
      \includegraphics{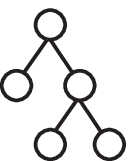}}
    &\scalebox{1}[1]{\includegraphics{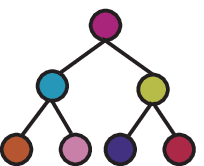}} \\
    (a) graph &(b) PAGGM \\ \\
    \multicolumn{2}{c}{
      \scalebox{1}[1]{\includegraphics{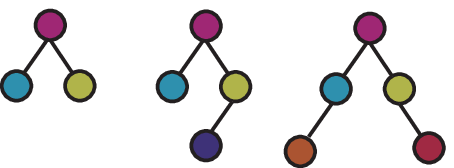}}} \\ 
    \multicolumn{2}{c}{(c) Median graphs} \\ \\
    \multicolumn{2}{c}{
      \scalebox{1}[1]{\includegraphics{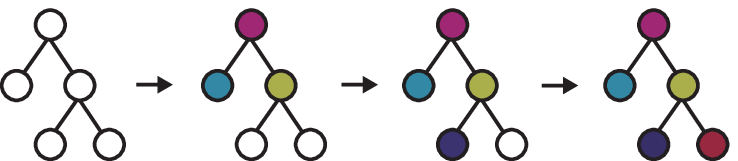}}}\\
    \multicolumn{2}{c}{(d) Process of assigning labels to the graph.}
  \end{tabular}
  \caption{
    Calculation of matching matrix between graph (a) and PAGGM (b).
    The colors represent labels. The correspondence develops from left to right in (c). 
  } \label{fig:matching}
\end{figure}

It is difficult to calculate $Y$, which maximizes Eq. (\ref{eq:prob}), when the number of vertices of the PAGGM is large. Unfortunately, the number of vertices is usually large. Therefore, we propose calculating $Y$ with a set of median graphs calculated in Algorithm 1. The number of vertices of the median graph is small and, most importantly, the vertices and edges of the median graphs correspond to the PAGGM. Hence, the correspondence between $G'$ and $P$ can be obtained through the median graphs. Specifically, we choose one of the median graphs and calculate a matching matrix between the median graph and $G'$. Then, we assign vertices in $G'$ to $P$ according to the matching matrix. We repeat this assignment by choosing another median graph until all vertices in $G'$ corresponding to the PAGGM or all median graphs have been chosen. Finally, we obtain matching matrix $Y$. The edit distance algorithm \cite{Neuhaus:2006:SSPR} is used to calculate the matching matrices between $G'$ and the median graphs. We illustrate an example of the calculation of $Y$ in Fig. \ref{fig:matching}. 

\section{Graph Model Boosting} \label{sec:Boosting}
\begin{figure*}[t] \centering
  \includegraphics[width=7.2in]{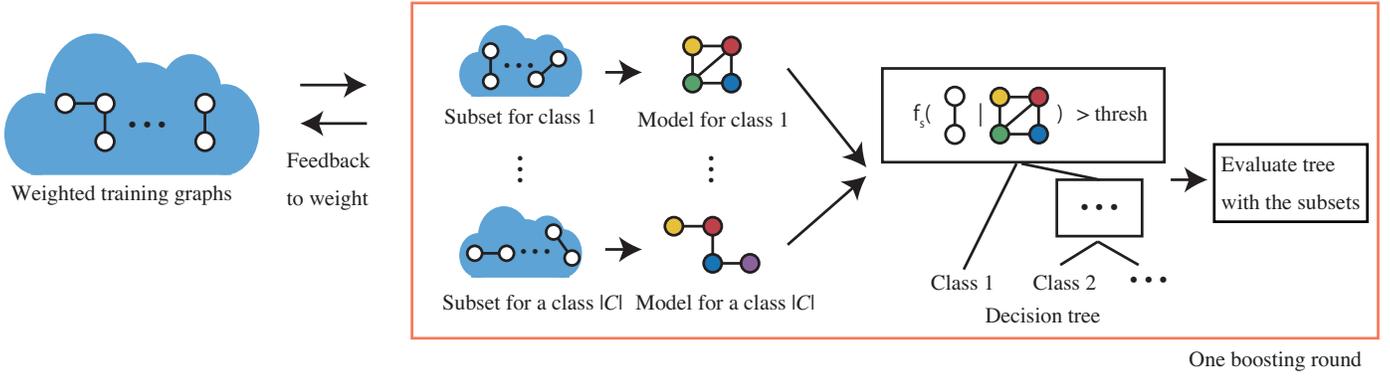}
  \caption{Overview of graph model boosting.} \label{fig:boosting}
\end{figure*}

The aim of GMB is to capture comprehensive variation
and construct a powerful classifier with a large number of graph models.
Although the PAGGM can contain structural variations,
the PAGGM is single and tends to capture major variations in the training graphs.
Hence, minor but important variations are lost.
Consequently, recognition errors occur on graphs in minor structures.
To overcome this problem, we adopt a boosting algorithm to generate a large number of PAGGMs
to complement one another. 
We provide an overview of the GMB in Fig. \ref{fig:boosting}.
In each boosting round, a PAGGM is constructed
with a subset of weighted training data for each class.
Note that the PAGGM can focus on minor graphs with large weights.
Then, we form a decision tree that classifies a graph
by recursively branching left or right down the tree according to the similarity
between the graph and a PAGGM.
Finally, we form a strong classifier by integrating the decision trees. 

\subsection{Model extension for weighted training data}
We extend the PAGGM for weighted training data. We begin by extending the median graph to a {\it weighted median graph}. Let $W$ be a set of weights for graphs in $\bm{G}$. We define the weighted median graph $\hat{M}$ as 
\begin{equation}
  \hat{M} = \argmin_{G \in \bm{G}} \sum_{G_i \in \bm{G}} w_i f_d( G, G_i ). 
\end{equation}
We replace the median graph in Algorithm 1 with the weighted median graph.
Therefore, $\bm{M}$ of $P$ is composed of $\hat{M}$.
Subsequently, we extend the calculation of $B$ and $\Theta$ to incorporate $W$ as
\begin{eqnarray}
  \hat{b}_i &=& \frac{1}{|\bm{G}|} \sum_{G_k \in \bm{G}} \sum_{v \in V_k} w_k
  \mathbb{I}\left( \mathcal{L}(v)=i \right), \\
  \hat{b}_{ij} &=& \frac{1}{|\bm{G}|} \sum_{G_k \in \bm{G}} \sum_{e_{v,v'} \in E_k}
  w_k \mathbb{I}\left( \{ \mathcal{L}(v), \mathcal{L}(v') \} = \{i,j\} \right), \\
  \hat{A}_i &=& \bigcup_{G_k \in \bm{G} }
  \bigcup_{\substack{v_l \in V_k\\ \mathcal{L}(v_l)=i}} a_l \diamond \mathcal{O}(w_l),\\
  \hat{A}_{ij} &=& \bigcup_{G_k \in \bm{G} }
  \bigcup_{\substack{e_{l,m} \in E_k\\ \{\mathcal{L}(v_l),\mathcal{L}(v_m)\} = \{i,j\}}}
  a_{l,m} \diamond \mathcal{O}(w_{l,m}),
\end{eqnarray}
where $\mathcal{O}$ is a function that refers to the position of $w_k$ in ascending order of $W$. For example, given $W = (3, 6, 1, 9)$, $\mathcal{O}(6)=3$.  Let $\diamond$ represent a duplication operator, such as $x \diamond N = \{ \underbrace{x, x, \cdots, x}_{N\mbox{ times}} \}$.  We calculate $\hat{\Theta}$ with $\hat{A}$. We replace $B$ and $\Theta$ with $\hat{B}$ and $\hat{\Theta}$, respectively, when $W$ is given.

\subsection{Boosting framework}
We propose GMB as shown in Algorithm 2,
where $C$ represents a set of class labels of $\bm{G}$.
GMB is based on the AdaBoost algorithms \cite{Freund:1997, SAMME, Allwein:2001:JMLR},
which are a suitable choice for model construction
because they provide different subsets of training data.
We construct PAGGMs using the subsets.
In addition to the different subsets, the weight can diversify the PAGGMs.
In the AdaBoost framework, the weights of error-prone graphs become larger than those of recognized graphs.
Hence, we can focus on such error-prone graphs by constructing PAGGMs using weights. 

\begin{figure}[t]
  \includegraphics{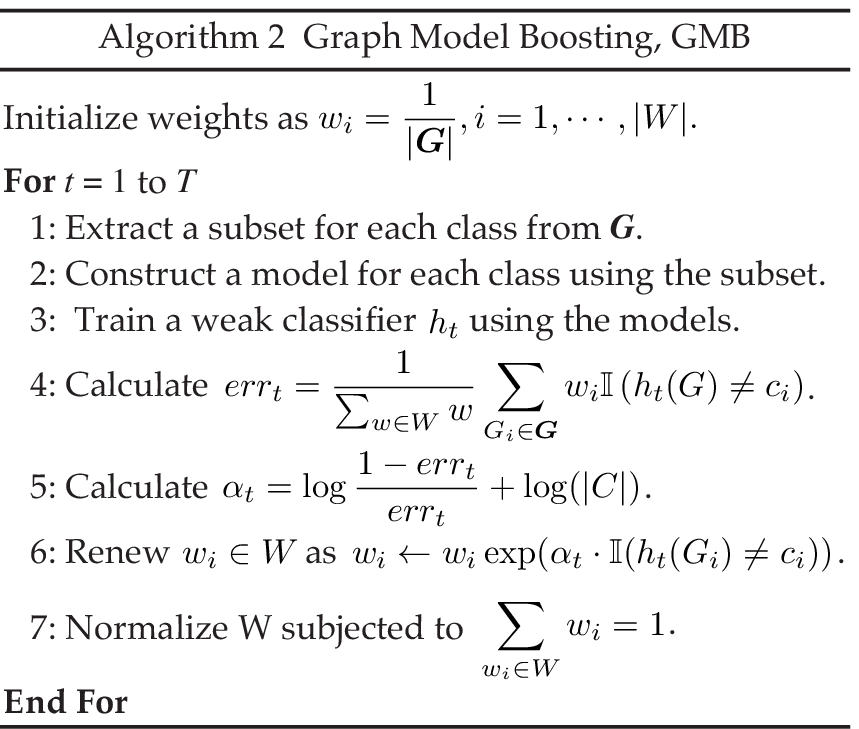}
\end{figure}

We use decision trees \cite{Breiman:1984} as weak classifiers $h_t: G \to c \in C$. 
The trees are trained by searching branching rules composed of the PAGGM of a class
and a threshold of similarity.
The trees have different PAGGMs and branching rules because of randomness in the GMB.
Subsequently, we use all trained trees to construct the strong classifier $H$ as 
\begin{equation} 
  H(G) = \argmax_{c \in C} \sum_{t=1}^T \alpha_t \mathbb{I}\left( h_t(G) = c \right). 
\end{equation}
In recognition, $H$ evaluates test graphs using the trees
and recognizes them by majority voting on the trees. 
Hence, the test graphs are comprehensively analyzed by measuring similarities to the PAGGMs
and using more sophisticated recognition rules than NN and k-NN.

\section{Experiments} \label{sec:experiments}
We carried out the experiments below to verify the effectiveness of the proposed method. We measured recognition accuracy for PAGGM and GMB on five datasets.

\subsection{Datasets}
For the experiments, we used five datasets from the IAM graph database repository \cite{IAM:Repository}, which is widely used for graph-based pattern recognition and machine learning. We used the following datasets: Letter-LOW (LOW), Letter-MED (MED), Letter-HIGH (HIGH), GREC, and COIL-RAG (COIL). 

The LOW, MED, and HIGH contained graphs representing distorted letters, which were the 15 capital letters of the Roman alphabet composed of straight lines only, such as A, E, F, H, I, K, L, M, N, T, U, V, W, X, Y, and Z.  The terms ``LOW,'' ``MED,'' and ``HIGH'' represent the levels of distortion. 

The GREC consisted of graphs representing symbols from architectural and electronic drawings. The symbols were converted into graphs by assigning vertices to corners, intersections, and circles on the symbols. The lines between symbols were edges. There were 22 classes in the GREC. As suggested in \cite{Sole:2012:SSPR}, we used graphs consisting of only straight lines. 

The COIL was composed of 100 classes of graphs extracted from images of different objects of the COIL-100 database \cite{COIL-100}. The images were segmented into regions according to color by the mean shift algorithm \cite{MeanShift}.  The vertices were assigned to the segmented regions, and adjacent regions were connected by edges. The attributes were color histograms. 

We summarize the characteristics of the datasets in Table \ref{hyo:Data}. We calculated $\Theta$ in the following probability density functions: two-dimensional normal distribution for attributes of (x,y), discrete distribution function for line type and RGB histogram, and one-dimensional normal density for the line boundary.

\begin{table*}[t] \centering
  \caption{
    Summary of characteristics of datasets used in the experiments.  The Top row indicates dataset name, the number of training data items, test data, classes, attribute types of vertex and edge, average numbers of vertices and edges, maximum number of vertices and edges, and the number of graphs in a class. 
  } \label{hyo:Data}
  \begin{tabular}{l|cccccccccc} 
    &\#train &\#test & \#class &vertex attribute &edge attribute
    &ave(v)  &ave(e)  &max(v)  &max(e) &\#a class\\ \hline
    LOW &750 &750 &15 &(x,y) & none &4.7 &3.1 &8 &6 &50\\ 
    MED &750 &750 &15 &(x,y) & none &4.7 &3.2 &9 &7 &50\\ 
    HIGH &750 &750 &15 &(x,y) & none &4.7 &4.5 &9 &9 &50\\ 
    GREC &213 &390 &22 &(x,y) & Line type &11.6 &11.9 &24 &26 &15\\
    COIL &2 400 &1 000 &100 &RGB histogram &Line boundary &3.0 &3.0 &11 &13 &24
  \end{tabular}
\end{table*}

\subsection{Experiments with PAGGM}
We carried out experiments with PAGGM.
In order to evaluate PAGGM, we constructed one PAGGM for each class by using all training data. We then classified the test data by applying the NN method with similarity obtained by Eq. (\ref{eq:similarity}). For comparison, we calculated the median graphs \cite{Jiang:2001:PAMI} and classified test data by the edit distance algorithm \cite{Neuhaus:2006:SSPR}. The experimental results are shown in Table \ref{hyo:model}. 

\begin{table}[t] \centering
  \caption{Recognition rates (\%) of median graph and PAGGM using the NN method.} \label{hyo:model}
  \begin{tabular}{l|ccccc}
    &LOW &MED &HIGH &GREC &COIL \\ \hline
    Median  &84.3 &77.3 &78.0 &84.4 &58.3 \\
    PAGGM &{\bf 95.2} &{\bf 87.5} &{\bf 85.9} &{\bf 90.0} &{\bf 71.1}
  \end{tabular}
\end{table}

The results show that the PAGGM outperformed the median graph for every dataset. Both the PAGGM and the median graph were categorized into the model-based approach. The main difference was whether it contained structural variation. These results signify that recognition performance improved when we incorporated structural variation. The experimental results showed that the PAGGM successfully captured structural variation.

\subsection{Experiments with Graph Model Boosting}
\begin{figure} \centering
  \includegraphics[width=3.2in]{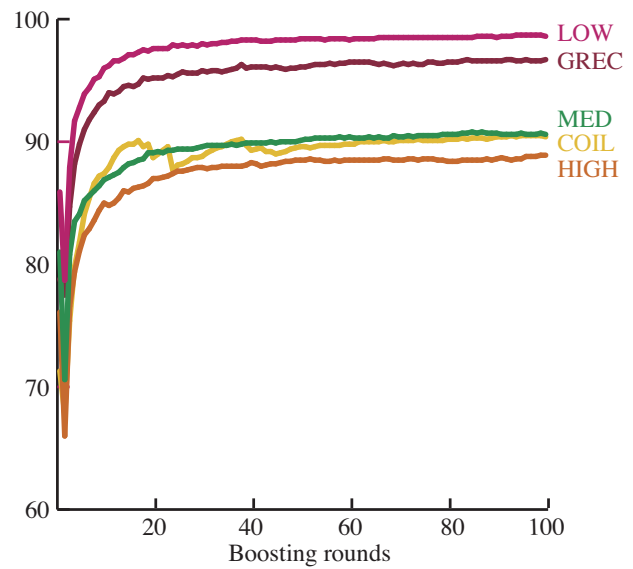}
  \caption{Evolutions of average recognition rates of GMB
    in $100$ boosting rounds.
    The evolutions mark the highest values as:
    $98.7\%$ at 94 round on LOW, $90.8\%$ at 85 round on MED, $88.9\%$ at 99 round on HIGH,
    $96.7\%$ at 84 round on GREC, and $91.9\%$ at 83 round on COIL.
    Best viewed in color.} \label{fig:evolution}
\end{figure}

We proved the effectiveness of GMB by showing the evolution of its recognition rate.
The recognition rate fluctuated during the trials due to random sampling.
We repeated the experiments 10 times and calculated the average recognition rate.
In this experiment, we set the cardinality of subset $n_c$ to 5 in LOW,
35 in MED, 40 in HIGH, 6 in GREC, and 24 in COIL,
where the number of boosting rounds $T$ was $100$.
Fig. \ref{fig:evolution} illustrates the evolutions.
The evolutions showed that the average recognition rates increased steadily on every dataset.
The initial average recognition rates were lower than the results with the single PAGGM,
shown in Table \ref{hyo:model}.
At the beginning, GMB only captured the structural variation in subsets of the training data,
resulting in poor results.
However, the accuracy of GMB increased as the process continued and,
finally, exceeded the results of the single PAGGM.
This phenomenon signifies that comprehensive structural variations were successfully incorporated
into the classifier of the GMB. 

\begin{figure*}[t] \begin{center}
    \includegraphics[width=6.9in]{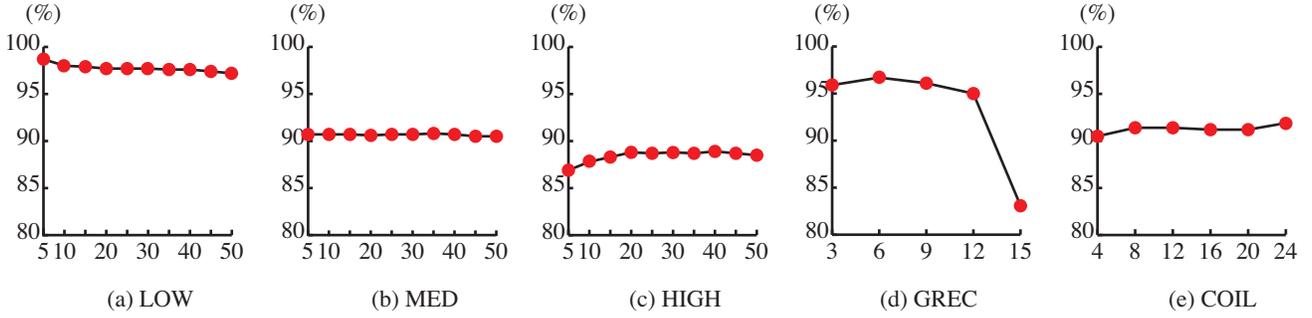}
    \caption{
      The effects of the cardinarity of a subset $n_c$.
      The y- and x-axes represent average recognition rate (\%) and $n_c$, respectively. 
      The highest recognition rates were as follows:
      (a) $98.7\%$ at $n_c = 5$, (b) $90.8\%$ at $n_c=35$,
      (c) $88.9\%$ at $n_c=40$, (d) $96.7\%$ at $n_c=6$, (e) $91.9\%$ at $n_c=24$. }
    \label{fig:AD_NUM}
\end{center} \end{figure*}

We carefully determined $n_c$ since random sampling is an important process.
We repeated this experiment 10 times for each $n_c$
and calculated the maximum values of the averages recognition rates.
Fig. \ref{fig:AD_NUM} shows the results.
The results show the robustness of GMB.
In general, the effects were within a few points of one another on the datasets.
GMB achieved good results on the datasets using various $n_c$.
Note that the results in LOW and GREC signified that performance can be maximized
when the number of samples is small.
The small number of samples facilitated the generation of different PAGGMs.
Consequently, more comprehensive structural variations were incorporated into the GMB.

Graph matching is an NP-complete problem,
and encounters a computation bottleneck in the GMB in Steps 2 and 4 in Algorithm 1.
Although we adopted a method to calculate approximate matching \cite{Neuhaus:2006:SSPR},
it took a considerable amount of time to perform over 100 rounds of boosting.
Therefore, to handle this computation problem,
we performed graph matching on all pairs prior to GMB.
In the GMB, we simply referred to the results.
The average processing times over 100 boosting rounds were 6.3 (s/round) in LOW,
12.9 (s/round) in MED, 8.6 (s/round) in HIGH, 13.5 (s/round) in GREC, and 631.4 (s/round) in COIL.
Note that the number of classes and the cardinality of the subset in COIL were greater than
in the other datasets. Consequently, the processing times were longer.

We compared GMB with existing graph recognition methods,
categorized into the one-vs-one, the model-based, and the embedding approaches.
We selected the following methods:
Riesen's method \cite{Riesen:2009:GbR} from the one-vs-one approach,
and Bunke's method \cite{Bunke:2012:PRL} from the embedding approach,
He's \cite{He:2006}, Wong's \cite{Wong:1985:PAMI}, Serratosa's \cite{Serratosa:2003:PR},
and Sanfeliu's \cite{SanFeLiu:2004} methods from the model-based approach.
The comparisons on LOW, HIGH, and GREC are summarized in Table \ref{hyo:Comparison}.
We referred to the results of
\cite{Riesen:2009:GbR,He:2006,Wong:1985:PAMI,Serratosa:2003:PR,SanFeLiu:2004}
to \cite{Sole:2012:SSPR} on the IAM database, whereas the original papers were not evaluated.
Although the methods in the one-vs-one and the embedding approaches recorded high scores,
the methods in the model-based approach obtained even higher scores.
These results signify the importance of structural information.
In the graph domain, sophisticated recognition rules need structural information
to perform to their full capabilities.
The PAGGM obtained high scores on every dataset.
Furthermore, GMB outperformed the other methods.
We successfully incorporated a large amount of structural information
into the powerful recognition rule.
These comparison results verified the effectiveness of GMB. 

\begin{table}[t] \centering
  \caption{Comparison of recognition rate with those of other methods.} \label{hyo:Comparison}
  \begin{tabular}{ll|ccc}
      Method &Approach &LOW &HIGH &GREC \\ \hline
      Riesen \cite{Riesen:2009:GbR} &One. &91.1 &77.6 &- \\ 
      Bunke \cite{Bunke:2012:PRL} &Emb. &92.9 &- &- \\ 
      Jiang \cite{Jiang:2001:PAMI} &Model &96.4 &74.5 &80.5 \\ 
      He \cite{He:2006} &Model &93.6 &49.1 &57.0 \\ 
      Wong \cite{Wong:1985:PAMI} &Model &93.9 &80.1 &85.8 \\ 
      Serratosa \cite{Serratosa:2003:PR} &Model &93.9 &80.1 &85.8 \\
      SanFeLiu \cite{SanFeLiu:2004} &Model &94.0 &80.9 &91.2 \\
      Proposed (PAGGM) & Model &95.2 &85.9 &90.0 \\
      Proposed (GMB) &Boosting &{\bf 98.7} &{\bf 88.9} &{\bf 96.7}
  \end{tabular}
\end{table}

Finally, to demonstrate the capability of GMB, we carried out further experiments 
whether GMB works well with other graph models.
To this end, we used the median graph as graph models constructed in GMB.
We show the results in Table \ref{tab:table_median}.
The recognition rates increased in all datasets.
These results signify the capability of GMB.

\begin{table}[t] \begin{center}
    \caption{Recognition rates (\%) of GMB using median graph as model.}
    \label{tab:table_median}
      \begin{tabular}{l|ccccc}
        &LOW &MED &HIGH &GREC &COIL \\ \hline
        Initial &91.3 &77.8 &67.8 &84.8 &76.3 \\
        Max &95.0 &88.8 &86.1 &94.1 &91.4
      \end{tabular}
\end{center} \end{table}

\section{Conclusions}
In this paper, we introduced PAGGM and GMB, a novel graph model and algorithm for graph recognition.
PAGGM captures structural variations in composition and attribute of vertex and edge.
We proposed an efficient algorithm for searching vertex correspondences
among multiple graphs.
The algorithm enables us to construct PAGGM in a feasible amount of time.
GMB constructs a large number of PAGGMs to comprehensively capture the structural variation of graphs.
Then, we formed a strong classifier using the constructed PAGGMs in the boosting framework.
Consequently, the classifier was equipped with the requisite information concerning structural variation
and a powerful recognition ability.

The experimental results showed that PAGGM successfully captured structural variation
and GMB significantly enhances recognition performance.
Furthermore, GMB outperformed the methods proposed in past work.
Therefore, we successfully strengthened the recognition capability and the ability
to deal with structural variation in graphs.

Structural data are a powerful representation of objects even in images. 
The proposed method can be applied to computer vision tasks
where relation of visual features needs to be considered.
Developing object recognition applications using the proposed method is planned for future work.

\ifCLASSOPTIONcompsoc
  \section*{Acknowledgments}
\else
  \section*{Acknowledgment}
\fi
This research was supported by JSPS KAKENHI Grant Number 15H06009.

\ifCLASSOPTIONcaptionsoff
  \newpage
\fi



%\bibliographystyle{IEEEtran}
%\begin{thebibliography}{1}
%H.~Kopka and P.~W. Daly, \emph{A Guide to {\LaTeX}}, 3rd~ed.\hskip 1em plus
%\end{thebibliography}
\bibliographystyle{IEEEtran}
\bibliography{refs}


\begin{IEEEbiography}[{\includegraphics[width=1in,height=1.25in,clip,keepaspectratio]{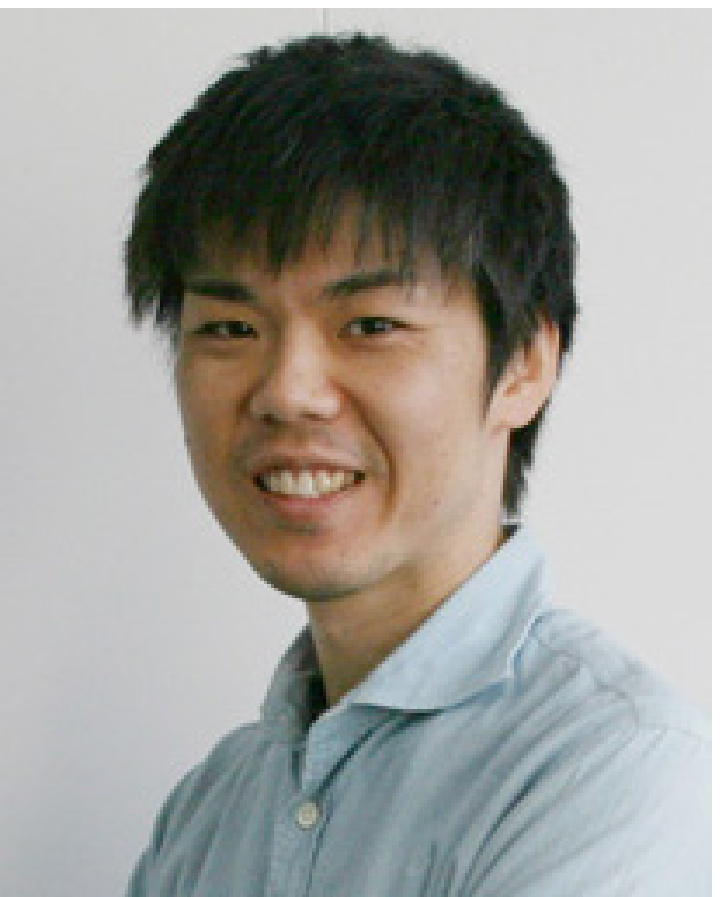}}]{Tomo Miyazaki}
  received a B.E. degree from the Department of Informatics,
  Faculty of Engineering, Yamagata University in 2006. He received M.E. and Ph.D. degrees
  from the Graduate School of Engineering, Tohoku University in 2008 and 2011, respectively.
  He joined Hitachi, Ltd. in 2011 and has worked at the Graduate School of Engineering,
  Tohoku University from 2013 to 2014 as a researcher.
  Since 2015, he has been an Assistant Professor.
  His research interests include pattern recognition and image processing.
  Dr. Miyazaki is a member of the Institute of Electronics,
  Information and Communication Engineers.
\end{IEEEbiography}

\begin{IEEEbiography}[{\includegraphics[width=1in,height=1.25in,clip,keepaspectratio]{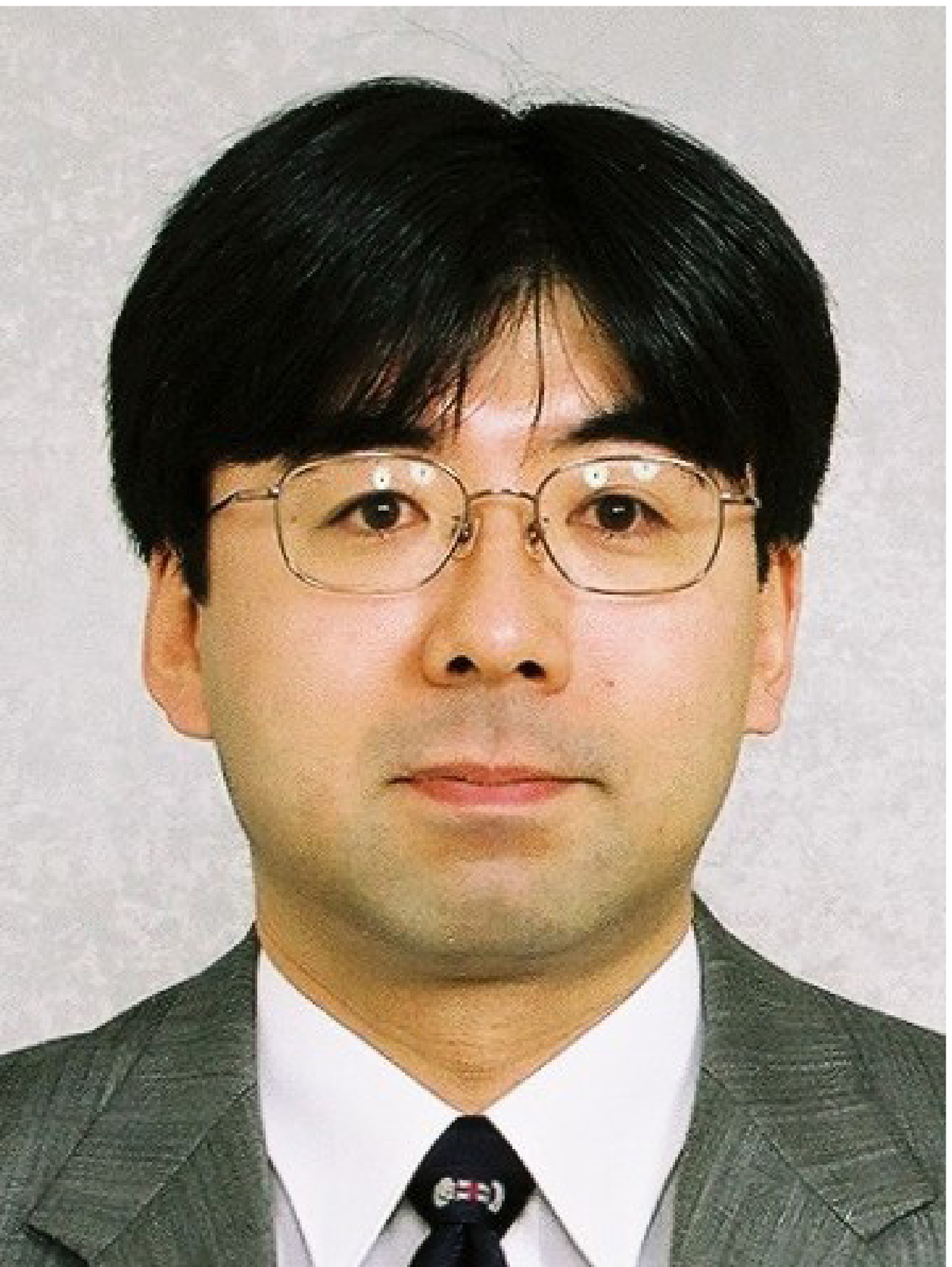}}]{Shinichiro Omachi}
  received B.E., M.E., and Doctor of Engineering degrees
  in Information Engineering from Tohoku University, Japan,
  in 1988, 1990, and 1993, respectively. He worked as a research associate
  at the Education Center for Information Processing at Tohoku University
  from 1993 to 1996. Since 1996, he has been with the Graduate School of Engineering at Tohoku University, where he is currently a professor. From 2000 to 2001, he was a visiting associate professor at Brown University. His research interests include pattern recognition, computer vision, image processing, image coding, and parallel processing. He received the IAPR/ICDAR Best Paper Award in 2007, the Best Paper Method Award of the 33rd Annual Conference of the GfKl in 2010, the ICFHR Best Paper Award in 2010, and the IEICE Best Paper Award in 2012. Dr. Omachi is a member of the Institute of Electronics, Information and Communication Engineers, among others.
\end{IEEEbiography}


%\newpage



%\vfill


\end{document}